\documentclass[a4paper, 10pt, conference]{ieeeconf}      
\usepackage{FG2025}
\usepackage{comment}
\usepackage{soul}
\usepackage{hyperref}
\FGfinalcopy 

\IEEEoverridecommandlockouts                              
\usepackage{amsmath} 
\usepackage{amssymb}  
\usepackage{caption} 
\usepackage{soul}
\usepackage[flushleft]{threeparttable}
\usepackage{array}
\usepackage{booktabs} 
\usepackage{multirow} 
\usepackage{adjustbox} 
\usepackage{amsmath} 
\usepackage{ulem} 
\usepackage{caption} 
\usepackage{graphics}
\usepackage{graphicx}
\usepackage{calc}

\def\FGPaperID{****} 

\title{\LARGE \bf Gender Fairness of Machine Learning Algorithms for Pain Detection}

\usepackage{fancyhdr}
\begin{document}


\author{
Dylan Green$^{*,1}$, Yuting Shang$^{*,1}$,
Jiaee Cheong$^{1,2}$, Yang Liu$^{3}$, Hatice Gunes$^{1}$\\
\thanks{
\noindent
*Both authors contributed equally to this work.
$^{1}$ Department of Computer Science and Technology, University of Cambridge. 
$^{2}$ Harvard University. 
$^{3}$ Center for Machine Vision and Signal Analysis, University of Oulu. 
}%
}

\maketitle

\thispagestyle{fancy} 

\begin{abstract}

Automated pain detection through machine learning (ML) and deep learning (DL) algorithms holds significant potential in healthcare, particularly for patients unable to self-report pain levels. However, the accuracy and fairness of these algorithms across different demographic groups (e.g., gender) remain under-researched. This paper investigates the gender fairness of ML and DL models trained on the UNBC-McMaster Shoulder Pain Expression Archive Database, evaluating the performance of various models in detecting pain based solely on the visual modality of participants' facial expressions. We compare traditional ML algorithms, Linear Support Vector Machine (L SVM) and Radial Basis Function SVM (RBF SVM), with DL methods, Convolutional Neural Network (CNN) and Vision Transformer (ViT), using a range of performance and fairness metrics. While ViT achieved the highest accuracy and a selection of fairness metrics, all models exhibited gender-based biases. These findings highlight the persistent trade-off between accuracy and fairness, emphasising the need for fairness-aware techniques to mitigate biases in automated healthcare systems.


\end{abstract}


\section{Introduction}

Machine Learning (ML) has become an essential tool in modern healthcare, offering the potential to automate complex tasks, such as pain detection, through images and videos \cite{Pain_Survey}. However, as these technologies are adopted, ensuring fairness becomes critical to avoid perpetuating or exacerbating existing biases \cite{bias, gender_bias_healthcare,spitale2024underneath}. 

ML fairness refers to the absence of prejudice or bias in a machine learning system concerning sensitive attributes such as gender, race, or age \cite{mehrabi2021survey}. In pain detection models, fairness ensures that individuals across different demographic groups are equally likely to be correctly classified. More specifically, gender fairness focuses on providing equal treatment and outcomes for male and female groups \cite{mehrabi2021survey, hitchhiker,cheong2023s,kwok2025machine}. For example, a fair pain detection system would ensure equal probability of predicting pain for individuals regardless of gender, assuming equivalent pain intensity \cite{hitchhiker}. 
While various fairness metrics such as Equalised Odds and Equal Accuracy exist
\cite{fairnessdefinitions, fairness_metric}, 
no single metric captures all aspects of fairness, making this a challenging but necessary area of research 
\cite{MLFairness}.

Bias in ML systems can originate from dataset bias in the input data, or from algorithmic bias in the models themselves \cite{mehrabi2021survey, sun2020evolution, bernhardt2022potential}. This study focuses on investigating algorithmic bias, aiming to address disparities introduced by ML models beyond those present in the dataset. Understanding and mitigating these biases is particularly important in healthcare, where biased systems can disproportionately harm underrepresented or historically marginalised groups.

Pain detection is a critical task in clinical settings, aiding healthcare professionals in monitoring patients and making informed decisions. Automated pain detection systems, particularly those based on facial expressions, are promising because they are non-invasive, real-time, and practical \cite{Pain_Survey}. 
Yet, existing research has primarily focused on improving accuracy without sufficient attention to fairness \cite{UNBC_Pain}. 
This gap is particularly concerning given documented gender disparities in healthcare. Studies show that societal and cultural norms influence how men and women perceive, express, and report pain, often leading to biases in clinical assessments
\cite{brave_men}. For instance, gendered stereotypes, such as associating masculinity with higher pain tolerance, can lead to underdiagnosis or misdiagnosis \cite{brave_men}.

The motivation for this study stems from the need to develop trustworthy ML systems that are both accurate and fair 
\cite{vashney2022trustworthy, varona2022discrimination}. 
Ensuring fairness is a fundamental aspect of trustworthiness, as unbiased systems are more likely to gain user confidence by delivering just outcomes \cite{feng2023review} and not discriminating. 
It is essential to scrutinize the decisions made by automated processes, since biases can exacerbate existing disparities in healthcare systems \cite{adhd, autism}. 
Utilising the UNBC dataset \cite{UNBC_Pain}, 
this study examines how algorithmic bias affects gender fairness in pain detection models, which is a research gap we have identified.
This research contributes to the field by implementing multiple unimodal pain detection classifiers and evaluating their performance through both accuracy and fairness metrics. The findings aim to provide a foundation for developing equitable pain detection systems and encourage further research into fairness across other healthcare applications and affective computing tasks.

\section{Literature Review}

\subsection{Methods for Pain Detection}

Pain detection 
employ a range of scales and techniques to quantify pain, including self-assessment tools, observer-based scales, and emerging automated pain detection systems.
Common self-reporting scales include the Visual Rating Scale (VRS), Visual Analog Scale (VAS), and Numeric Rating Scale (NRS) \cite{lopez2017personalized, werner2019automatic}. 
The VRS allows patients to describe pain intensity qualitatively using categories like "mild", "moderate", or "severe", offering a straightforward and intuitive measure \cite{Jensen2003PainScales}. 
Similarly, VAS provides a continuous scale in which patients mark their perceived pain intensity on a line, delivering finer granularity than categorical methods \cite{Hawker2011PainAssessment}. 
The NRS simplifies quantification by asking patients to rate their pain level on a numerical scale, typically 0 (no pain) to 10 (worst imaginable pain) \cite{Breivik2008NumericalRating}.

For patients unable to self-report, healthcare professionals can use scales like the Critical-Care Pain Observation Tool (CPOT) and Behavioural Pain Scale (BPS) to assess patients based on observed behavioural indicators \cite{observer-pain-scale}. The Prkachin and Solomon Pain Intensity (PSPI) metric is a specialised observational tool that quantifies pain intensity through facial action units (AUs). This objective metric has become particularly significant in computer vision research, as demonstrated in datasets such as the UNBC-McMaster Shoulder Pain Archive \cite{UNBC_Pain}.

ML approaches for automated pain detection range from traditional methods like logistic regression and SVMs to deep learning architectures including CNNs \cite{logreg-paindetection,ml-paindetection,dl-paindetection}. While 
Gkikas \textit{et al.}
\cite{dl-paindetection} demonstrates the value of multimodal approaches, most studies have not adequately addressed fairness and robustness across diverse populations.


\subsection{Bias and Fairness in Affective Computing}


Fairness studies in affective computing are crucial for developing equitable classification systems, as models often exhibit biases due to unrepresentative training data.
Till date, most of the existing literature on bias and fairness in affective computing have chiefly focused on facial expression recognition \cite{hitchhiker,cheong2023counterfactual,cheong2023causal}
and depression detection 
\cite{cheong2023towards,cheong2024fairrefuse,u-fair_ml4h_2024}
\sloppy{In Facial Expression Recognition (FER), systems frequently demonstrate lower performance on underrepresented demographic groups, primarily due to biased datasets \cite{xu2020investigating}. 
Addressing these issues requires data balancing and the application of fairness-aware algorithms. Similar challenges are observed in Action Unit (AU) detection, where disparities in performance across gender and ethnicity have been documented \cite{churamani2022domain}. These findings emphasise the importance of diverse training data for enhancing model equity.
}

Depression prediction models exhibit disparities that may indicate gender and cultural biases, highlighting the necessity for culturally adaptive systems \cite{Multimodal_Fairness,cheong2025fairness}. Similarly, studies on postpartum depression detection emphasise the value of gender-sensitive models \cite{saqib2021machine, escriba2011gender}. These challenges in psychological state modelling reveal broader issues in affective computing. Together, these works reinforce the pressing need to address biases in affective computing tasks to ensure fair and robust outcomes.
Moreover, till date, none of the existing work has investigated ML bias and fairness in pain detection.


\subsection{Comparison of ML and DL Approaches}


The comparative performance of traditional Machine Learning (ML) and Deep Learning (DL) methods in pain detection reveals significant trade-offs, with implications for fairness, interpretability, and performance.
Traditional ML methods, such as Support Vector Machines (SVMs), excel in settings with smaller datasets and handcrafted features, often delivering robust performance with lower computational requirements \cite{Scholkopf2002SVM}. 
In contrast, DL models leverage large datasets to capture complex patterns, offering superior accuracy in certain contexts but requiring extensive computational resources and careful dataset curation \cite{LeCun2015DeepLearning, CNN_Depression}.
%
ML models, due to their simpler decision boundaries, may exhibit lower bias compared to DL models \cite{wang2023bias}. 
Meanwhile, DL models can amplify biases inherent in the training data \cite{mehrabi2021survey}, which can be addressed through fairness-aware training techniques and diverse datasets \cite{Barocas2019FairnessML}. Moreover, interpretability is a notable advantage of traditional ML approaches, facilitating bias detection and correction \cite{Doshi2017Interpretability}. On the other hand, DL models often function as "black boxes", complicating fairness evaluations and rectifications of any biases \cite{black_box}. These comparisons stress the need for systematic evaluations of ML and DL approaches, particularly in applications demanding fairness and transparency.

\subsection{Research Gaps and Challenges}

Despite advancements in pain detection and affective computing, several critical research gaps (RGs) remain unaddressed, particularly concerning fairness and methodological comparisons.

\begin{itemize}
    \item \textbf{RG 1:} Current research on fairness largely focuses on other affective computing tasks such as FER and AU detection, with limited exploration in examining demographic biases in pain detection.
    
    \item \textbf{RG 2:} Additionally, existing datasets, including the UNBC-McMaster database, suffer from insufficient demographic representation, constraining the development of equitable models \cite{wang2023bias}. 
    
    \item \textbf{RG 3:} Although various ML and DL methods have been applied to pain and depression detection, few studies compare multiple models within a fairness context, especially for gender fairness within pain detection. Current research, like the study by \cite{Multimodal_Fairness}, emphasises the importance of this for depression. However, gender-specific fairness in the area of ML pain detection has been under-explored.
    
\end{itemize}

Addressing these gaps will improve the reliability and clinical applicability of automated pain detection systems, making them fairer across gender groups and enhancing their usefulness in diverse healthcare settings. 
%
\section{Experimental Setup}

Our experiments utilise the UNBC dataset \cite{UNBC_Pain}.
%
The dataset is split into training, validation, and testing sets, ensuring balanced representation across genders and pain levels. 
We train and evaluate:
\begin{itemize}
    \item Two ML models, Linear Support Vector Machine (SVM) and Radial Basis Function Support Vector Machine (RBF SVM)
    \item Two DL models, namely Convolutional Neural Network (CNN) and Vision Transformers (ViT) 
\end{itemize}
%
focusing on their performance with a variety of fairness measures (Table~\ref{tab:metrics}). 
This setup provides a foundation for detailed analysis in the subsequent sections.

\subsection{Dataset \& Pain measurements}

The UNBC dataset consists of 200 sequences from 25 different subjects with shoulder pain, totalling 48,398 frames. 
Each frame has been Facial Action Coding System (FACS) coded, where each facial expression is described in terms of AUs \cite{UNBC_Pain}. 
This complements the PSPI metric from which we can quantify pain from facial expression data. For the UNBC dataset, the number of frames with the associated PSPI score is given in Table \ref{tab:pspi}. The resulting PSPI score for a frame is then calculated using the following equation:
\begin{equation}
    \text{Pain} = \text{AU}4 + \text{Max}(\text{AU}6, \text{AU}7) + \text{Max}(\text{AU}9, \text{AU}10) + \text{AU}43
\end{equation}
%
Within the UNBC dataset, 
PSPI ranges on a scale between $0-16$ 
where 0 represents ``no-pain" 
and 16 represents the highest pain intensity \cite{prkachin2008structure}. 
Within our binary classification setting, any images labelled with a PSPI score greater than 0 are labelled as the target class ``pain" ($Y=1$)
%
and images 
with $PSPI=0$ are labelled as ``no-pain". 
%
The distribution of the PSPI scores in the UNBC dataset is shown in Table~\ref{tab:pspi}. 
We see that 
the majority of frames being labelled with no pain ($Y=0$) and only about one-sixth of the dataset labelled as with pain ($Y=1$).

\captionsetup{justification=centering}
\begin{table}[ht]
    \centering
    \caption{Distribution of Pain Intensity Scores (PSPI) in the UNBC Dataset}
    \label{tab:pspi}
    \begin{tabular}{c|c}
        \textbf{PSPI Score} & \textbf{Frequency}\\
        \hline
        0 & 40029\\
        1-2 & 5260\\
        3-4 & 2214\\
        5-6 & 512\\
        7-8 & 132\\
        9-10 & 99\\
        11-12 & 124\\
        13-14 & 23\\
        15-16 & 5\\
    \end{tabular}
\end{table}

In the UNBC dataset, there are 25 participants, 13 female and 12 male. 
The images were manually labelled and independently cross-verified by two researchers for gender.
and we used stratified sampling to ensure a balanced data split across the training, validation, and testing datasets for each participant. This approach also ensured that both pain and no-pain images for each participant were included, so that the gender fairness evaluation of our models was not influenced by the distribution of pain-related images. The split was done 60/20/20 in accordance with the suggested best practice \cite{muraina2022ideal} across the image dataset as shown in Table~\ref{tab:datasplit}.

\captionsetup{justification=centering}
\begin{table}[ht]
    \centering
    \caption{UNBC Data Split Before Dataset Balancing}
    \label{tab:datasplit}
    \begin{tabular}{c|ccc}
        & \textbf{Training} & \textbf{Validation} & \textbf{Testing} \\
        \hline
         Pain & 5012 & 1673 & 1684\\  
        No pain & 24007 & 8006 & 8016\\
    \end{tabular}
\end{table}

\subsection{pre-processing \& Data Augmentation}
\label{section:pre-processing}

After splitting the UNBC dataset, we focused on pre-processing the training dataset. As shown in Table~\ref{tab:datasplit}, there is a large disparity between the pain and no-pain classes, with the no-pain data being almost 5 times more than the pain data \cite{UNBC_Pain}. 
This imbalance is likely due to pain typically being triggered in short episodes. 
Unbalanced datasets can lead to poor accuracy in pain classification and introduce dataset biases that models may learn \cite{hitchhiker}. 
Since our goal is to investigate whether ML models introduce algorithmic biases, it is critical to first address any biases present in the dataset. 
%
Oversampling the minority pain class from 5012 to 24007 samples is a good approach to improve the imbalance ratio, whilst also increasing the number of training samples available
\cite{imbalance, balance_trick}.

\subsubsection{SMOTE}
\begin{figure}[ht]
    \includegraphics[width=2.5cm, height=2.5cm]{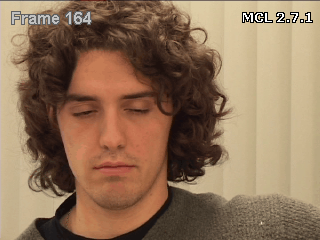}
    \hspace{0.2cm}
    \includegraphics[width=2.5cm, height=2.5cm]{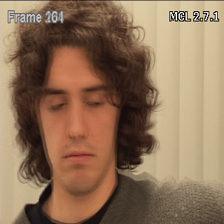}
    \hspace{0.2cm}
    \includegraphics[width=2.5cm, height=2.5cm]{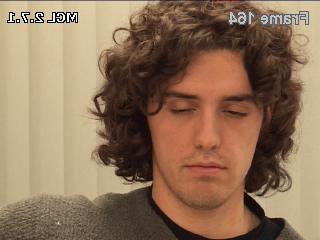}

    \vspace{0.2cm} 

    \includegraphics[width=2.5cm, height=2.5cm]{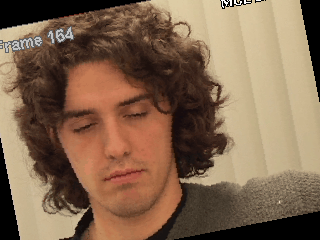}
    \hspace{0.2cm}
    \includegraphics[width=2.5cm, height=2.5cm]{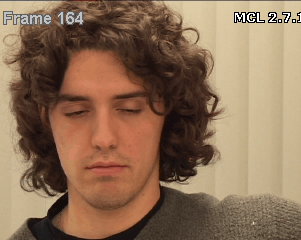}
    \hspace{0.2cm}
    \includegraphics[width=2.5cm, height=2.5cm]{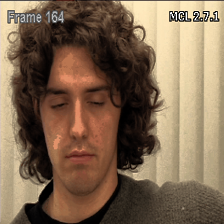}

    \caption{From left to right, the images represent: (1) Original, (2) SMOTE, (3) Flipped, (4) Rotated, (5) Cropped, and (6) Histogram Equalisation versions. All images are resized to squares for consistency in this diagram.}
    \label{fig:smote_output}
\end{figure}

The first step in pre-processing involved using the Synthetic Minority Oversampling Technique (SMOTE) to increase the pain class dataset by a factor of 4.75 times \cite{balance_trick}. SMOTE generates synthetic samples for the minority class by interpolating existing data points, which mitigates the overfitting risk of simple duplication methods \cite{smote, to_smote}. 
Oversampling was performed separately for each participant to prevent generating unrealistic data points that would mix characteristics from different people. 
We applied \texttt{SMOTE} from \texttt{imblearn.over\_sampling} with \texttt{k\_neighbors} set to 5 (this describes the number of nearest samples used for synthesis) \cite{scikit-learn}. Image pixel arrays were input into SMOTE, and the output was reconstructed back into pixel values for a \(224 \times 224\) image. An example output (2) is shown in Figure~\ref{fig:smote_output}: the frame number in the top left corner is visibly distorted from the interpolation process.

The pain-class dataset was increased to 23,798 samples through SMOTE. To equalise the number of pain and no-pain samples, a small subset of pain-class images was randomly duplicated, bringing both classes to 24,007 samples.

\subsubsection{Data augmentation}
To enhance variation in the training data and improve robustness against overfitting, we performed data augmentation on both pain and no-pain classes \cite{hitchhiker}. Four augmentation techniques were used: flipping (along the left-right axis), rotation (by a random angle \(-20\leq \theta \leq 20\) \cite{rotation}), cropping (a random 0-5\% from the left and right sides, avoiding the central facial region), and histogram equalisation (adjusting the intensity distribution of the HSV value channel) \cite{intensity_hist_equ, hsv}. Each transformation was applied to a random fifth of the pain and no-pain images, with the possibility of multiple transformations being applied to the same image. Examples of transformations (3)--(6) can be seen in Figure~\ref{fig:smote_output}.

To provide consistent training datasets for both the ML and DL models, the pre-processing steps were controlled by setting a random seed of 42 across all experiments 
%
Post-pre-processing, four random images were visually cross-verified by researchers to confirm identical transformations.

\begin{table*}[ht!]
\centering
\caption{Evaluation Metrics and their Formulae}
\label{tab:metrics}
\begin{threeparttable}
    \centering
\begin{tabular}{>{\centering\arraybackslash}p{2.5cm}p{9cm}c}

\textbf{Metric}      & \textbf{Description} & \textbf{Formula}  \\

\toprule
\addlinespace[0.5em] 

Accuracy     & The ratio between the correctly predicted images against all the images in the dataset.                         & 
\begin{minipage}[c]{5cm} 
\begin{equation}
\frac{\text{TP} + \text{TN}}{\text{TP} + \text{TN} + \text{FP} + \text{FN}}
\label{eq:accuracy}
\end{equation}
\end{minipage}
\\

F1 Score     & The harmonic mean of precision and recall\tnote{*} , weighted by class distribution, balancing false positives (misclassified as pain) and false negatives (misclassified as no-pain).                        & 
\begin{minipage}[c]{5cm} 

\begin{equation}
2 \times \frac{\text{Precision} \times \text{Recall}}{\text{Precision} + \text{Recall}}
\label{eq:f1}
\end{equation}
\end{minipage}
\\

ROC AUC     & The ability of a model to distinguish between pain and no-pain classes, with values closer to 1 indicating better performance. The ROC curve is a plot of TPR against FPR\tnote{**}, and AUC is the area under that curve.                       & 
\begin{minipage}[c]{5cm} 
\vspace{-\abovedisplayskip}
\begin{equation}
\int^1_0\text{TPR } d\text{FPR}
\label{eq:roc}
\end{equation}
\end{minipage}
\\

\midrule
\addlinespace[0.5em] 

Equal Accuracy     & Measures if male and female groups have equal rates of accuracy. & 
\begin{minipage}[c]{5cm} 
\begin{equation}
\text{Accuracy}_{\text{Male}}
 = \text{Accuracy}_{\text{Female}}
\label{eq:equalacc}
\end{equation}
\end{minipage}
\\

Equal Opportunity    & Measures if male and female groups have equal TPRs. \tnote{**} & 
\begin{minipage}[c]{5cm} 
\begin{equation}
\text{TPR}_{\text{Male}} = \text{TPR}_{\text{Female}}
\label{eq:equalopp}
\end{equation}
\end{minipage}
\\

Equalised Odds     & Measures if male and female groups have equal TPR and FPR, where the actual pain label \(Y\) is 1 or 0 respectively\tnote{**}. & 
\begin{minipage}[c]{5cm} 
\begin{equation}
\mathbb{P}(\hat{Y}=1\mid Y, \text{Male}) = \mathbb{P}(\hat{Y}=1\mid Y, \text{Female})
\label{eq:eo}
\end{equation}
\end{minipage}
\\

Disparate Impact     & Compares the ratio of minority over majority individuals that receive a positive pain outcome of 1, where \(\hat{Y}\) is the predicted value.             & 
\begin{minipage}[c]{5cm} 
\begin{equation}
\frac{\mathbb{P}(\hat{Y} = 1 \mid \text{Male})}{\mathbb{P}(\hat{Y} = 1 \mid \text{Female})}
\label{eq:disimp}
\end{equation}
\end{minipage}
\\

Demographic (Statistical) Parity     & Measures if subjects in both groups have equal probability of being
assigned to the positive pain class 1, where \(\hat{Y}\) is the predicted value. & 
\begin{minipage}[c]{5cm} 
\begin{equation}
\mathbb{P}(\hat{Y}=1 \mid \text{Male}) = \mathbb{P}(\hat{Y}=1 \mid \text{Female})
\label{eq:dempar}
\end{equation}
\end{minipage}
\\

Treatment Equality     & Measures if male and female groups have the same ratio of FNs to FPs. The range of this metric is not bound between 0 and 1.& 
\begin{minipage}[c]{5cm} 
\begin{equation}
\frac{\text{FN}_{\text{Male}}}{\text{FP}_{\text{Male}}} = \frac{\text{FN}_{\text{Female}}}{\text{FP}_{\text{Female}}}
\label{eq:te}
\end{equation}
\end{minipage}
\\

Test Fairness     & For any predicted value \(\hat{Y}\), subjects in both groups have equal probability of truly belonging to the positive pain class, where \(Y\) is the actual value. & 
\begin{minipage}[c]{5cm} 
\begin{equation}
\mathbb{P}(Y=1\mid\hat{Y}, \text{Male}) = \mathbb{P}(Y=1\mid\hat{Y}, \text{Female})
\label{eq:tf}
\end{equation}
\end{minipage}
\\

\bottomrule
\addlinespace[0.5em] 

\end{tabular}
\tnote{*}{where:
$
\text{Precision} = \frac{TP}{TP + FP} \quad \text{and} \quad 
\text{Recall} = \frac{TP}{TP + FN} \quad
$}
\\
\tnote{**}{where:
$
\text{TPR}_{\text{Group}} = \frac{TP_{\text{Group}}}{TP_{\text{Group}} + FN_{\text{Group}}}= \mathbb{P}(\hat{Y}=1\mid Y=1)  \quad \text{and} \quad 
\text{FPR}_{\text{Group}} = \frac{FP_{\text{Group}}}{FP_{\text{Group}} + TN_{\text{Group}}} = \mathbb{P}(\hat{Y}=1 \mid Y=0) 
$}
\end{threeparttable}

\end{table*}

\subsection{Classical ML Models Tuning \& Training Implementation}

In the context of classical ML, we investigated the SVM model, using two different kernel functions: 
\begin{itemize}
    \item the Linear kernel and
    \item the Radial Basis Function (RBF) kernel \cite{svm_survey}. 
\end{itemize}

Intuitively, SVMs separate data points into two classes by constructing a flat decision plane (a hyperplane) between the data points, and maximising the margin between the two classes of data \cite{svm_survey}. 
Linear SVMs are suitable for image classification tasks since the dimension of the input feature space (e.g. image pixels) is large, which often leads to linearly separable data \cite{svm_img_cls}. 
For data that cannot be linearly separated, one can use a non-linear SVM. 
This applies a non-linear kernel transformation (e.g. RBF), to elevate the data points to a higher dimensionality, in order to find an optimal separating hyperplane (OSP) \cite{svm_img_cls}.


\subsubsection{Linear SVM} The first model explored was the Linear SVM. In order to enhance practicality on a large dataset, the classifier was constructed using \texttt{SGDClassifier} from \texttt{sklearn.linear\_model}, with the default hinge loss and L2 penalty (regularisation) parameters \cite{scikit-learn,sgd_svm}. Stochastic Gradient Descent (SGD) serves as an optimisation technique, where a single random sample is selected to update the loss gradient at each iteration \cite{sgd_cls}.

\begin{figure}[ht]
     \includegraphics[width=2.5cm, height=2.5cm]{Images/a1.png} 
    \hspace{0.2cm} 
    \includegraphics[width=2.5cm]{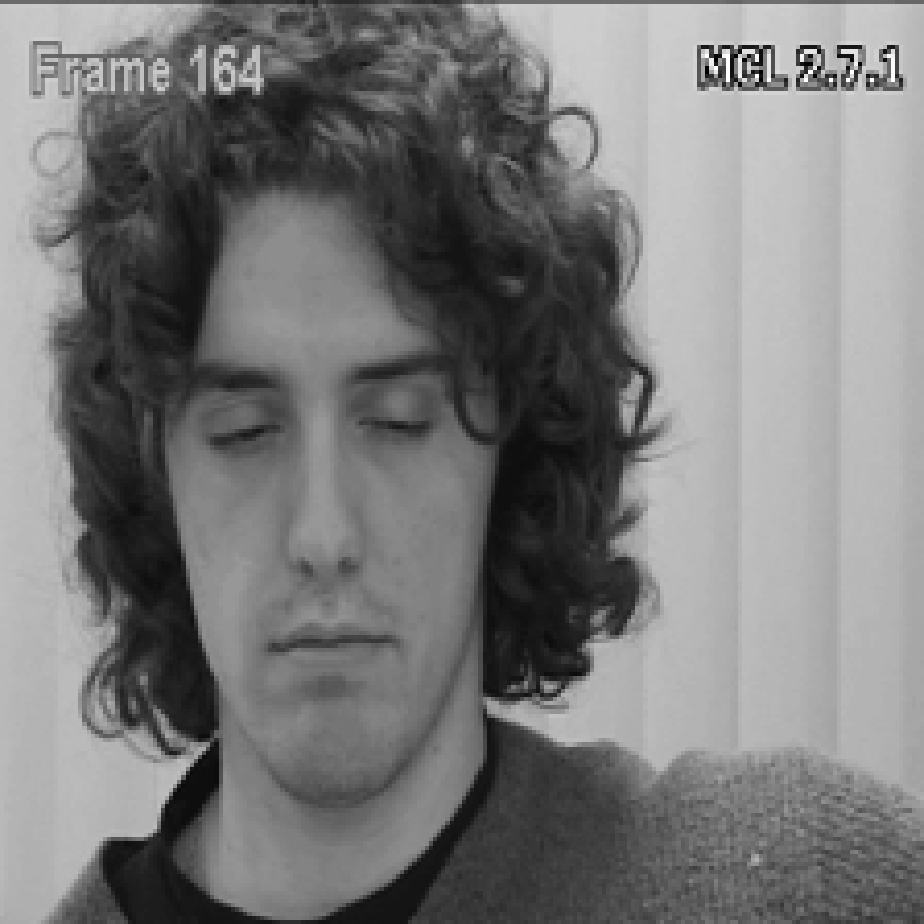} 
    \hspace{0.2cm} 
    \includegraphics[width=2.5cm]{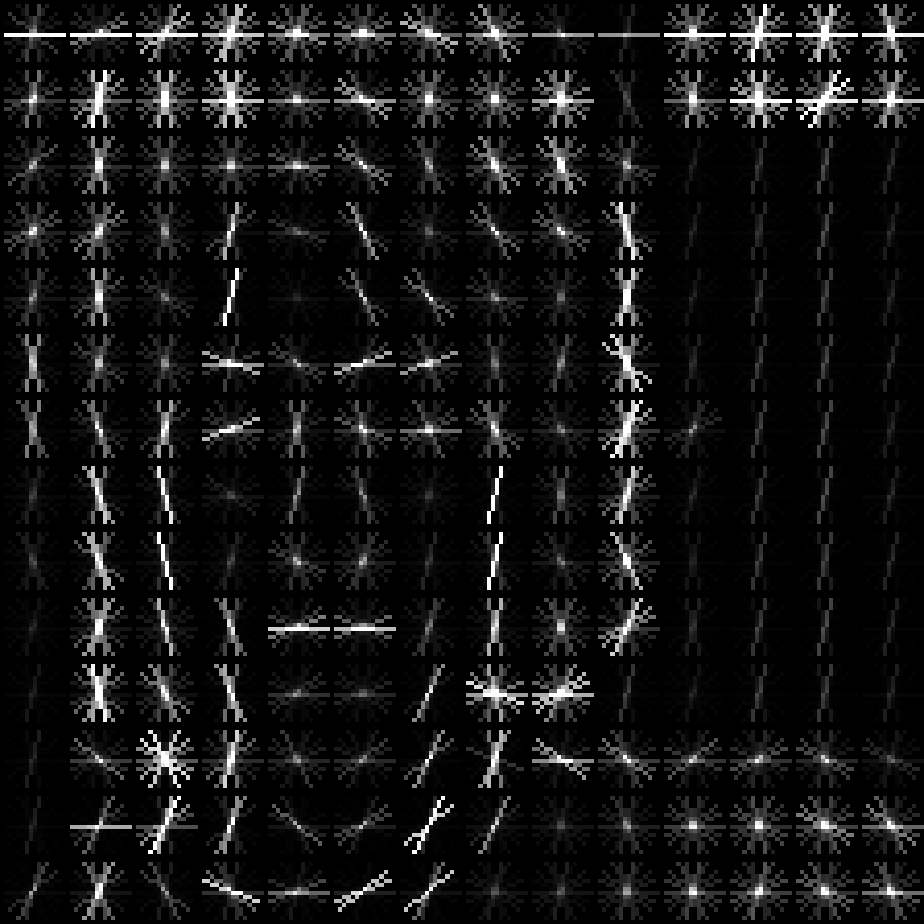} 
    \caption{HOG feature extraction. From left to right, the images represent (1) Original, (2) Grayscale, (3) HOG features}
    \label{fig:hog_extraction}
\end{figure}

\vspace{1mm}
%
\noindent\textbf{Pre-processing:}
To prepare the dataset for training, pre-processing steps as outlined in Section~\ref{section:pre-processing} were first applied. 
Images were then resized to a standard resolution of \(224 \times 224\), converted to greyscale, then Histogram of Oriented Gradient (HOG) features were extracted for each image (as shown in Figure~\ref{fig:hog_extraction}) to reduce the input dimensionality of 150,528 \cite{hog_svm}. 
The cell size was set to \(16 \times 16\) pixels, and the block size to \(2 \times 2\), resulting in 6,084 features (this leads to a good sample-to-feature size ratio for each class \((N/L)\) of over three \cite{n_l_ratio}). 
Images were then standardised to a mean of 0 and standard deviation of 1, since SVM algorithms are sensitive to feature scaling \cite{scikit-learn}.

\vspace{1mm}
\noindent\textbf{Hyperparameter tuning:}
Hyperparameter tuning was performed using \texttt{GridSearchCV} from \texttt{sklearn.model\_selection} \cite{scikit-learn}. This process was initialised with a predefined split of the training and validation data, without cross-validation. The scoring function maximised accuracy, and tested each combination of hyperparameters supplied. The parameters tuned included alpha (the constant multiplier for regularisation), average (whether SGD weights are averaged), eta0 (the initial learning rate), and the type of learning\_rate \cite{scikit-learn}.

\vspace{1mm}
\noindent\textbf{Training and Implementation:}
The training process for the Linear SVM involved fitting the SGD classifier with the selected parameters: an alpha of 0.001, False for the averaging, an eta of 0.0001, and an adaptive learning rate.

\subsubsection{RBF SVM} The second model investigated was the RBF SVM. For practical implementation, the \texttt{Nystroem} transformer from \texttt{sklearn.kernel\_approximation} was utilised as a kernel approximation for RBF\cite{nystroem}. The combination of Nystr\"{o}m with the SGD classifier, enables non-linear learning \cite{scikit-learn}.

The same pre-processing steps applied to the training dataset for the Linear SVM were also used here, including HOG feature extraction and image vector standardisation. Hyperparameter tuning for both Nystr\"{o}m and the SGD classifier was conducted using the same grid search approach, scored based on accuracy. The parameters tuned include gamma (which controls the influence of a single sample in RBF), \texttt{n\_components} (the number of features to construct for Nystr\"{o}m
 approximation), and alpha \cite{scikit-learn}.

Training of the RBF SVM was carried out by constructing a pipeline that combined Nystr\"{o}m and the SGD classifier. The pipeline was then fitted with a gamma of 0.001, \texttt{n\_components} of 5000, and an alpha of 1e-07.



\subsection{DL Models Tuning \& Training Implementation}

We employed two deep learning models for our image classification task: a CNN based on the ResNet-50 architecture \cite{he2015deep} and a ViT model \cite{dosovitskiy2020image}.

\subsubsection{CNN (ResNet-50)}

ResNet-50 is a deep residual network with 50 layers that utilises skip connections to mitigate the vanishing gradient problem, enabling the efficient training of deep networks \cite{ResNetCNN} \cite{he2015deep}.

\vspace{1mm}
\noindent\textbf{Pre-processing:}
The pre-processing described in Section~\ref{section:pre-processing} was applied to the dataset and then the images were normalised using the mean and standard deviation of the ImageNet dataset, and converted to tensors \cite{ResNetCNN}. The ResNet-50 model was initialised with pre-trained weights from the ImageNet dataset, leveraging transfer learning to enhance performance with limited training data. The last few layers were unfrozen for fine-tuning, while the earlier layers were kept frozen to retain features learned during pre-training \cite{ResNetCNN}. The final fully connected layer was modified to output two classes for the binary classification problem that is pain detection.

\vspace{1mm}
\noindent\textbf{Hyperparameter tuning:}
Hyperparameter tuning was conducted using a grid search approach with the \texttt{ParameterGrid} from \texttt{sklearn.model\_selection} \cite{parametergrid} \cite{scikit-learn}. The hyperparameters tuned included learning rate, weight decay, batch size, and gradient accumulation steps. For each hyperparameter combination, the model was trained and validated, and the best configuration was selected based on validation loss.
%
The best configuration was then used in our subsequent training and implementation.

\vspace{1mm}
\noindent\textbf{Training and Implementation:}
The training process utilised the Adam optimiser with a learning rate of 0.0001 and a weight decay of 0.1. The cross-entropy loss function was used as the training criterion. Mixed precision training with gradient scaling enhanced computational efficiency and reduced memory usage \cite{micikevicius2018mixed}. Gradient accumulation effectively increased the batch size, with gradients accumulated over four mini-batches before an optimisation step. The model was trained with a batch size of 32, four accumulation steps, and 11 epochs. 

\subsubsection{Vision Transformer}

The ViT we used was specifically the \texttt{vit\_base\_patch16\_224} architecture \cite{ViTvsCNN}. The ViT segments an image into patches, linearly embeds each patch, and processes the sequence of embedded patches using a standard Transformer encoder. This architecture allows the model to capture long-range dependencies and complex patterns in the image data \cite{dosovitskiy2020image}.

\vspace{1mm}
\noindent\textbf{Pre-processing:}
The pre-processing applied to the dataset was identical to that used for the ResNet-50 model. The ViT model was also initialised with pre-trained weights from the ImageNet dataset, leveraging transfer learning to improve performance with limited training data \cite{dosovitskiy2020image}. 

\vspace{1mm}
\noindent\textbf{Hyperparameter Tuning:}
Unlike the ResNet-50 model, the entire ViT model was fine-tuned, enabling updates to all layers during training.
Hyperparameter tuning followed the same grid search approach as used with the CNN model. 
The best configuration was then used in our subsequent training and implementation.

\vspace{1mm}
\noindent\textbf{Training and Implementation:}
The training process utilised the Adam optimiser with a learning rate of 0.0001 and a weight decay of 0.1 \cite{loshchilov2017decoupled}. The cross-entropy loss function served as the training criterion. Mixed precision training with gradient scaling was employed to enhance computational efficiency \cite{micikevicius2018mixed}. Gradient accumulation effectively increased the batch size, with gradients accumulated over two mini-batches before performing an optimisation step. The ViT model was trained with a batch size of 16, two accumulation steps, and 10 epochs.
%

\subsection{Performance \& Fairness Evaluation}

After training, the overall performance and gender fairness of each model were evaluated using the metrics summarised in Table~\ref{tab:metrics}. The performance metrics include standard measures of Accuracy \eqref{eq:accuracy}, F1 Score \eqref{eq:f1}, 
and the Receiver Operating Characteristic Area Under the Curve
(ROC AUC) \eqref{eq:roc} 
\cite{performancemetrics, rocauc, auc_eq}, 
while the fairness metrics ~\eqref{eq:equalacc}--\eqref{eq:tf} aim to assess how well the models perform across the different gender groups\cite{hitchhiker, fairness_metric, stat_par}. For most fairness metrics, a value closer to $0.00$ is fairer, except for Disparate Impact \cite{disparateimpact}, where a value closer to $1.00$ is considered fairer.

\section{Experimental Results}

The evaluation results of our four models using the metrics from Table~\ref{tab:metrics} are presented in Table~\ref{tab:model_results}. Additionally, confusion matrices for the models, evaluated separately for males and females, are shown in Table~\ref{tab:confusion}. 

The test dataset consists of 4817 male samples and 4883 female samples. In the positive pain class, there are 553 male images and 1132 female images. The confusion matrix reveals gender-based disparities in both correct and incorrect classifications. However, to facilitate a more meaningful analysis, we will focus on the metrics in Table~~\ref{tab:metrics} rather than using the raw TP/FP/TN/FN values, 
as the latter lack context without appropriate ratios or perspective.

\subsection{Discussion and Findings}

This section discusses the implications of the findings, their alignment with existing literature, and potential avenues for improving fairness in ML systems.

\begin{table}[ht]
    \centering
    \caption{Model Performance \& Gender Fairness Results. Best values for each metric are in \textbf{bold}, worst in 
    \underline{underline}. Upward score trend ↑, downwards trend arrow omitted.
    Upward score trend means the higher the value the better performance or fairness (Accuracy/F1/ROC AUC/Disparate Impact). If there is no arrow, this means there is a downward score trend (i.e. the smaller the value/the closer the value to 0 the better)}
    \label{tab:results}
    \begin{tabular}{l|>{\centering\arraybackslash}p{\widthof{RBF SVM}}>{\centering\arraybackslash}p{\widthof{RBF SVM}}>{\centering\arraybackslash}p{\widthof{RBF SVM}}c}
        \toprule
        \textbf{Metric} & \textbf{CNN} & \textbf{ViT} & \textbf{L SVM} & \textbf{RBF SVM} \\
        \midrule
        Overall Acc {\tiny↑} & 0.9637 & \textbf{0.9806} & \underline{0.9509} & 0.9712 \\
        F1 Score {\tiny↑} & 0.9639 & \textbf{0.9808} & \underline{0.9514} & 0.9707 \\
        ROC AUC {\tiny↑} & 0.9902 & \textbf{0.9969} & \underline{0.9234} & 0.9331 \\
        \midrule
        Equal Acc & \underline{0.0098} & 0.0039 & 0.0051 & \textbf{0.0027} \\
        Equal Opp & 0.0801 & \textbf{0.0268} & 0.0763 & \underline{0.0908} \\
        Equal Odds & \underline{0.0488} & \textbf{0.0169} & 0.0423 & 0.0472 \\
        Dis Impact {\tiny↑} & 0.4650 & 0.4987 & \textbf{0.5099} & \underline{0.4572} \\
        Demo Parity & \underline{0.1300} & 0.1208 & \textbf{0.1175} & 0.1180 \\
        Tr Equality & 0.5564 & 0.1406 & \textbf{0.0030} & \underline{0.7825} \\
        Test Fairness & \textbf{0.0118} & 0.0161 & \underline{0.0512} & 0.0142 \\
        \bottomrule
    \end{tabular}
    \label{tab:model_results}
\end{table}

\subsubsection{Performance Evaluation}
Across all four models evaluated, ViT consistently achieved the highest performance metrics, with an overall accuracy \eqref{eq:accuracy} of 0.9806, an F1 score \eqref{eq:f1} of 0.9808, and a ROC AUC \eqref{eq:roc} of 0.9969. These results affirm the capability of ViT in capturing nuanced features from facial data, which aligns with prior studies highlighting its strength in image-based tasks \cite{dosovitskiy2020image}. 

In contrast, the Linear SVM exhibited the lowest performance across all metrics, achieving an overall accuracy \eqref{eq:accuracy} of 0.9509, an F1 score \eqref{eq:f1} of 0.9514 and a ROC AUC \eqref{eq:roc} of 0.9234. This demonstrates the limitations of linear models in handling complex patterns in high-dimensional data, while the RBF SVM, capable of modelling non-linearly separable data, outperformed the Linear SVM as expected \cite{svm_survey, svm_vs_cnn}. However, the RBF SVM outperforming the CNN in Accuracy ~\eqref{eq:accuracy} and F1 score \eqref{eq:f1} presents an interesting result. While other studies suggest that CNNs typically achieve better accuracy than classical ML models \cite{svm_vs_cnn}, 
the variation here could stem from differences in hyperparameter tuning or the relatively small UNBC dataset, which can be more suited for classical ML methods
\cite{ML_DL_health}. 

Both the CNN and ViT achieved high ROC AUC scores of 0.9902 and 0.9969, respectively, contrasting with the linear and RBF SVMs which had lower scores of 0.9234 and 0.9331. These results suggest that DL models have strong class-separating capabilities, 
however it is important to note debates regarding the reliability of ROC AUC as a performance metric in binary classification
\cite{roc_bad}.
%
However it is important to note debates regarding the reliability of ROC AUC as a performance metric in binary classification. ROC AUC only considers two of the confusion matrix basic rates, that is the TPR (true positive rate) and TNR (true negative rate). This could lead to overoptimistic inflated interpretations of the results when the pain classifier performs badly in terms of PPV (positive predictive value) or NPV (negative predictive value). Since the entire confusion matrix is provided in this paper, this can also be used in conjunction with the ROC AUC for a more informed overview of the results. However, metrics such as the Matthews Correlation Coefficient (MCC) could be used in future work, since it maximises all four of the confusion matrix basic rates, providing a more comprehensive summary of the classifier performance.
%


\begin{table}[!ht]
    \centering
    \caption{Confusion Matrices for Male and Female Splits for Each Model}
    \label{tab:confusion}
     \begin{tabular}{cc|>{\centering\arraybackslash}p{\widthof{RBF SVM}}>{\centering\arraybackslash}p{\widthof{RBF SVM}}>{\centering\arraybackslash}p{\widthof{RBF SVM}}c}
        \toprule
        \multicolumn{2}{c|}{\textbf{Gender}} & \textbf{CNN} & \textbf{ViT} & \textbf{L SVM} & \textbf{RBF SVM} \\
        \midrule
        \multirow{4}{*}{\rotatebox{90}{\textbf{Male}}} & TP & 473 & 524 & 459 & 450 \\
                                       & FP & 71  & 55  & 130 & 29  \\
                                       & TN & 4193 & 4209 & 4134 & 4235 \\
                                       & FN & 80  & 29  & 94  & 103 \\
        \midrule
        \multirow{4}{*}{\rotatebox{90}{\textbf{Female}}} & TP & 1058 & 1102 & 1025 & 1023 \\
                                         & FP & 128  & 75  & 146 & 39  \\
                                         & TN & 3624 & 3677 & 3606 & 3713 \\
                                         & FN & 73   & 29  & 106 & 108 \\
        \bottomrule
    \end{tabular}
\end{table}


\subsubsection{Fairness Evaluation}
Fairness metrics reveal notable disparities in gender-based treatment, even among the best-performing models. While ViT accomplished the best Equal Opportunity \eqref{eq:equalopp} (0.0268), and Equalised Odds \eqref{eq:eo} (0.0169), suggesting relative fairness of TPR and FPR across groups, it did not attain the best fairness in other metrics.

Linear SVM exhibited the lowest Demographic Parity \eqref{eq:dempar} of 0.1175, which is associated with having the highest Disparate Impact ratio \eqref{eq:disimp} of 0.5099. Both metrics suggest that the Linear SVM is relatively fairer in assigning the positive pain class, regardless of gender. This could imply that while simpler ML models tend to exhibit greater fairness compared to DL models \cite{wang2023bias}, they often sacrifice accuracy, illustrating the fairness-accuracy tradeoff \cite{fairness_tradeoff}. However, given the higher proportion of pain instances in the female test data compared to male pain images, these metrics \eqref{eq:disimp} \eqref{eq:dempar} do not provide conclusive evidence of algorithmic bias in the models \cite{kleinberg2016inherent}.

While the RBF SVM achieved the best Equal Accuracy \eqref{eq:equalacc} of 0.0027, it underperformed in Treatment Equality ~\eqref{eq:te} with a result of 0.7825. Treatment Equality is calculated from FN and FP values and is not bounded between 0 and 1 like other ratios, which may distort the interpretation of results. The high value for RBF SVM reveals a higher disparity in the FN/FP ratio between the two genders.

The CNN performed the best in Test Fairness ~\eqref{eq:tf} with a result of 0.0118, indicating that its predictions had similar relationships to true pain levels regardless of gender. However, CNNs exhibited the worst Equal Accuracy ~\eqref{eq:equalacc} of 0.0098, Equalised Odds ~\eqref{eq:eo} of 0.0488 and Demographic Parity ~\eqref{eq:dempar} of 0.1300. These results underscore the complexity of fairness analysis compared to performance analysis, as no model excels across all metrics. 
Recent studies have actually revealed that it is impossible to satisfy all fairness metrics simultaneously, a notion reflected in our results table 
\cite{MLFairness}. 
When working with smaller and imbalanced datasets, metrics such as Disparate Impact ~\eqref{eq:disimp} and Demographic Parity ~\eqref{eq:dempar} may be sensitive to group sizes \cite{kleinberg2016inherent}, while Equalised Odds ~\eqref{eq:eo} and Equal Opportunity ~\eqref{eq:equalopp} may better align with supervised learning objectives, encouraging features to be collected independent of protected attributes \cite{hardt2016equality}.

\section{Discussion and Conclusion}

This study provides critical insights into the trade-offs between performance and fairness in ML models for pain detection. 
Our evaluation identified gender biases across all models, with varying results across fairness metrics, highlighting that achieving gender equity remains an unresolved challenge. 
However, it is important to note that the ViT demonstrated both superior accuracy and consistent fairness measures in our results, being the only model to avoid worst-case performance across all fairness metrics. 
This could suggest that ViT's architecture, with its pre-training on diverse datasets and ability to capture global relationships, is less likely to overfit to biases in the dataset \cite{dosovitskiy2020image, hendrycks2020pretrained, hendrycks2021many}, offering a promising direction for mitigating biases in pain detection systems using advanced architectures. 

Several limitations warrant discussion. First, while this study employed commonly used fairness metrics, these may not capture all dimensions of bias, particularly in intersectional contexts involving multiple sensitive attributes (e.g., race and gender) \cite{fairness_metric,cheong2024small}. 
Second, the small and limited UNBC dataset may have constrained the generalisability of the findings \cite{UNBC_Pain}. Additionally, gender bias was introduced into the dataset during pre-processing while attempting to balance the class sizes. SMOTE oversampling increased gender imbalance (20,630 male vs. 27,384 female images) despite initial balanced distribution (14,413 male vs. 14,606 female). 

Further research is needed to explore post-hoc fairness interventions, such as adversarial de-biasing \cite{adversarial-debiasing}, fairness-aware loss functions \cite{fairness-aware}, 
uncertainty-based \cite{kuzucu2024uncertainty} or continual-learning \cite{churamani2023towards,cheong2024causal} based methods to enhance the fairness of pain detection systems without compromising accuracy. 
Additionally, the observed difficulty in distinguishing algorithmic bias from dataset bias suggests the need for more sophisticated data balancing techniques that maintain both class and demographic parity. These findings indicate the need for continued research into fairness of ML models and techniques to mitigate biases, ensuring that automated healthcare systems benefit all individuals equitably.

\section*{ETHICAL IMPACT STATEMENT}
\noindent
Our work focuses on advancing ethical ML and AI for the task of facial affect analysis systems.
We have adopted experimental measures aligned with ethical guidelines.
The datasets used have been anonymised to minimise privacy impact. Our work attempts to avoid any bias against certain groups of people that could result in discrimination. However, our results are limited to the dataset included in this work. 
Future work should repeat the same analysis on other pain-related datasets to further validate our findings. 
Overall, this research contributes to ethical ML and AI by raising awareness on the problem of ML bias in pain detection and encouraging best practices for fairer FAA systems.


\section*{ACKNOWLEDGMENTS}
\noindent
\textbf{Open Access:} For open access purposes, the authors have applied a Creative Commons Attribution (CC BY) licence to any Author Accepted Manuscript version arising.
\textbf{Data access:} This study involved secondary analyses of existing datasets. All datasets are described and cited accordingly. 
\textbf{Funding:} J. Cheong and H. Gunes have been supported by the EPSRC/UKRI under grant ref. EP/R030782/1 (ARoEQ) and EP/R511675/1. 
J. Cheong further acknowledges support from the Leverhulme Trust and the Wellcome Trust. The work of Dr. Yang Liu was supported in part by the Finnish Cultural Foundation for North Ostrobothnia Regional Fund under Grant 60231712, and in part by the Instrumentarium Foundation under Grant 240016. 


\bibliographystyle{ieee}
\bibliography{references,jiaee}

\begin{thebibliography}{10}\itemsep=-1pt

\bibitem{black_box}
A.~Aggarwal, P.~Lohia, S.~Nagar, K.~Dey, and D.~Saha.
\newblock Black box fairness testing of machine learning models.
\newblock In {\em Proceedings of the 2019 27th ACM joint meeting on european software engineering conference and symposium on the foundations of software engineering}, pages 625--635, 2019.

\bibitem{adhd}
D.~E. Attoe and E.~A. Climie.
\newblock Miss. diagnosis: a systematic review of adhd in adult women.
\newblock {\em Journal of attention disorders}, 27(7):645--657, 2023.

\bibitem{Barocas2019FairnessML}
S.~Barocas, M.~Hardt, and A.~Narayanan.
\newblock {\em Fairness and Machine Learning: Limitations and Opportunities}.
\newblock FairnessML, 2019.

\bibitem{bernhardt2022potential}
M.~Bernhardt, C.~Jones, and B.~Glocker.
\newblock Potential sources of dataset bias complicate investigation of underdiagnosis by machine learning algorithms.
\newblock {\em Nature Medicine}, 28(6):1157--1158, 2022.

\bibitem{Breivik2008NumericalRating}
H.~Breivik, P.-C. Borchgrevink, S.-M. Allen, L.-A. Rosseland, L.~Romundstad, E.~Breivik~Hals, G.~Kvarstein, and A.~Stubhaug.
\newblock Assessment of pain.
\newblock {\em British journal of anaesthesia}, 101(1), 2008.

\bibitem{fairness_tradeoff}
S.~Buijsman.
\newblock Navigating fairness measures and trade-offs.
\newblock {\em AI and Ethics}, 2023.

\bibitem{Multimodal_Fairness}
J.~Cameron, J.~Cheong, M.~Spitale, and H.~Gunes.
\newblock Multimodal gender fairness in depression prediction: Insights on data from the usa \& china.
\newblock {\em arXiv}, 08 2024.

\bibitem{fairness_metric}
A.~Castelnovo, R.~Crupi, G.~Greco, D.~Regoli, I.~G. Penco, and A.~C. Cosentini.
\newblock The zoo of fairness metrics in machine learning.
\newblock 2021.

\bibitem{gender_bias_healthcare}
S.~Caton and C.~Haas.
\newblock Fairness in machine learning: A survey.
\newblock {\em ACM Computing Surveys}, 55(3):1--35, 2023.

\bibitem{svm_vs_cnn}
S.~Y. Chaganti, I.~Nanda, K.~R. Pandi, T.~G. Prudhvith, and N.~Kumar.
\newblock Image classification using svm and cnn.
\newblock In {\em 2020 International Conference on Computer Science, Engineering and Applications (ICCSEA)}, pages 1--5, 2020.

\bibitem{svm_survey}
M.~A. Chandra and S.~Bedi.
\newblock Survey on svm and their application in image classification.
\newblock {\em International Journal of Information Technology}, 13(5):1--11, 2021.

\bibitem{svm_img_cls}
O.~Chapelle, P.~Haffner, and V.~N. Vapnik.
\newblock Support vector machines for histogram-based image classification.
\newblock {\em IEEE transactions on Neural Networks}, 10(5):1055--1064, 1999.

\bibitem{cheong2025fairness}
J.~Cheong.
\newblock {\em Fairness for affective and wellbeing computing}.
\newblock PhD thesis, 2025.

\bibitem{u-fair_ml4h_2024}
J.~Cheong, A.~Bangar, S.~Kalkan, and H.~Gunes.
\newblock U-fair: Uncertainty-based multimodal multitask learning for fairer depression detection.
\newblock In {\em Proceedings of the 4th Machine Learning for Health Symposium}, volume 259 of {\em Proceedings of Machine Learning Research}, pages 203--218. PMLR, 15--16 Dec 2025.

\bibitem{cheong2024causal}
J.~Cheong, N.~Churamani, L.~Guerdan, T.~E. Lee, Z.~Han, and H.~Gunes.
\newblock Causal-hri: Causal learning for human-robot interaction.
\newblock In {\em Companion of the 2024 ACM/IEEE International Conference on Human-Robot Interaction}, pages 1311--1313, 2024.

\bibitem{hitchhiker}
J.~Cheong, S.~Kalkan, and H.~Gunes.
\newblock The hitchhiker’s guide to bias and fairness in facial affective signal processing: Overview and techniques.
\newblock {\em IEEE Signal Processing Magazine}, 38(6):39--49, 2021.

\bibitem{cheong2023causal}
J.~Cheong, S.~Kalkan, and H.~Gunes.
\newblock Causal structure learning of bias for fair affect recognition.
\newblock In {\em Proceedings of the IEEE/CVF Winter Conference on Applications of Computer Vision}, pages 340--349, 2023.

\bibitem{cheong2023counterfactual}
J.~Cheong, S.~{Kalkan}, and H.~Gunes.
\newblock Counterfactual fairness for facial expression recognition.
\newblock In {\em ECCV 2022 Workshops}, pages 245--261. Springer, 2023.

\bibitem{cheong2024fairrefuse}
J.~Cheong, S.~Kalkan, and H.~Gunes.
\newblock Fairrefuse: referee-guided fusion for multimodal causal fairness in depression detection.
\newblock In {\em Proceedings of the Thirty-Third International Joint Conference on Artificial Intelligence}, pages 7224--7232, 2024.

\bibitem{cheong2023towards}
J.~Cheong, S.~Kuzucu, S.~Kalkan, and H.~Gunes.
\newblock Towards gender fairness for mental health prediction.
\newblock In {\em Proceedings of the Thirty-Second International Joint Conference on Artificial Intelligence}, pages 5932--5940, 2023.

\bibitem{cheong2023s}
J.~Cheong, M.~Spitale, and H.~Gunes.
\newblock “it’s not fair!”--fairness for a small dataset of multi-modal dyadic mental well-being coaching.
\newblock In {\em 2023 11th International Conference on Affective Computing and Intelligent Interaction (ACII)}, pages 1--8. IEEE, 2023.

\bibitem{cheong2024small}
J.~Cheong, M.~Spitale, and H.~Gunes.
\newblock Small but fair! fairness for multimodal human-human and robot-human mental wellbeing coaching.
\newblock {\em arXiv preprint arXiv:2407.01562}, 2024.

\bibitem{performancemetrics}
D.~Chicco and G.~Jurman.
\newblock The advantages of the matthews correlation coefficient (mcc) over f1 score and accuracy in binary classification evaluation.
\newblock {\em BMC genomics}, 21:1--13, 2020.

\bibitem{roc_bad}
D.~Chicco and G.~Jurman.
\newblock The matthews correlation coefficient (mcc) should replace the roc auc as the standard metric for assessing binary classification.
\newblock {\em BioData Mining}, 16(1):4, 2023.

\bibitem{churamani2023towards}
N.~Churamani, J.~Cheong, S.~Kalkan, and H.~Gunes.
\newblock Towards causal replay for knowledge rehearsal in continual learning.
\newblock In {\em AAAI Bridge Program on Continual Causality}, pages 63--70. PMLR, 2023.

\bibitem{churamani2022domain}
N.~Churamani, O.~Kara, and H.~Gunes.
\newblock Domain-incremental continual learning for mitigating bias in facial expression and action unit recognition.
\newblock {\em IEEE Transactions on Affective Computing}, 14(4):3191--3206, 2022.

\bibitem{hog_svm}
H.~S. Dadi and G.~M. Pillutla.
\newblock Improved face recognition rate using hog features and svm classifier.
\newblock {\em IOSR Journal of Electronics and Communication Engineering}, 11(04):34--44, 2016.

\bibitem{Doshi2017Interpretability}
F.~Doshi-Velez and B.~Kim.
\newblock Towards a rigorous science of interpretable machine learning.
\newblock {\em arXiv preprint arXiv:1702.08608}, 2017.

\bibitem{dosovitskiy2020image}
A.~Dosovitskiy, L.~Beyer, A.~Kolesnikov, D.~Weissenborn, X.~Zhai, T.~Unterthiner, M.~Dehghani, M.~Minderer, G.~Heigold, S.~Gelly, J.~Uszkoreit, and N.~Houlsby.
\newblock An image is worth 16x16 words: Transformers for image recognition at scale.
\newblock {\em arXiv preprint arXiv:2010.11929}, 2020.

\bibitem{to_smote}
Y.~Elor and H.~Averbuch-Elor.
\newblock To smote, or not to smote?
\newblock {\em arXiv preprint arXiv:2201.08528}, 2022.

\bibitem{smote}
D.~Elreedy and A.~F. Atiya.
\newblock A comprehensive analysis of synthetic minority oversampling technique (smote) for handling class imbalance.
\newblock {\em Information Sciences}, 505:32--64, 2019.

\bibitem{escriba2011gender}
V.~Escrib{\`a}-Ag{\"u}ir and L.~Artazcoz.
\newblock Gender differences in postpartum depression: a longitudinal cohort study.
\newblock {\em Journal of Epidemiology \& Community Health}, 65(4):320--326, 2011.

\bibitem{feng2023review}
T.~Feng, R.~Hebbar, N.~Mehlman, X.~Shi, A.~Kommineni, S.~Narayanan, et~al.
\newblock A review of speech-centric trustworthy machine learning: Privacy, safety, and fairness.
\newblock {\em APSIPA Transactions on Signal and Information Processing}, 12(3), 2023.

\bibitem{n_l_ratio}
D.~Foley.
\newblock Considerations of sample and feature size.
\newblock {\em IEEE Transactions on Information Theory}, 18(5):618--626, 1972.

\bibitem{ML_DL_health}
V.~C. Gandhi and P.~P. Gandhi.
\newblock A survey-insights of ml and dl in health domain.
\newblock In {\em 2022 International Conference on Sustainable Computing and Data Communication Systems (ICSCDS)}, pages 239--246. IEEE, 2022.

\bibitem{stat_par}
P.~Garg, J.~Villasenor, and V.~Foggo.
\newblock Fairness metrics: A comparative analysis.
\newblock In {\em 2020 IEEE international conference on big data (Big Data)}, pages 3662--3666. IEEE, 2020.

\bibitem{dl-paindetection}
S.~Gkikas and M.~Tsiknakis.
\newblock Automatic assessment of pain based on deep learning methods: A systematic review.
\newblock {\em Computer methods and programs in biomedicine}, 231:107365, 2023.

\bibitem{hardt2016equality}
M.~Hardt, E.~Price, and N.~Srebro.
\newblock Equality of opportunity in supervised learning.
\newblock {\em Advances in neural information processing systems}, 29, 2016.

\bibitem{Pain_Survey}
T.~Hassan, D.~Seus, J.~Wollenberg, K.~Weitz, M.~Kunz, S.~Lautenbacher, J.~U. Garbas, and U.~Schmid.
\newblock Automatic detection of pain from facial expressions: A survey.
\newblock {\em IEEE Transactions on Pattern Analysis and Machine Intelligence}, 43(6):1815--1831, 2021.

\bibitem{Hawker2011PainAssessment}
G.~A. Hawker, S.~Mian, T.~Kendzerska, and M.~French.
\newblock Measures of adult pain: Visual analog scale for pain (vas pain), numeric rating scale for pain (nrs pain), mcgill pain questionnaire (mpq), short-form mcgill pain questionnaire (sf-mpq), chronic pain grade scale (cpgs), short form-36 bodily pain scale (sf-36 bps), and measure of intermittent and constant osteoarthritis pain (icoap).
\newblock {\em Arthritis care \& research}, 63(S11):S240--S252, 2011.

\bibitem{he2015deep}
K.~He, X.~Zhang, S.~Ren, and J.~Sun.
\newblock Deep residual learning for image recognition.
\newblock In {\em Proceedings of the IEEE conference on computer vision and pattern recognition}, pages 770--778, 2016.

\bibitem{hendrycks2021many}
D.~Hendrycks, S.~Basart, N.~Mu, S.~Kadavath, F.~Wang, E.~Dorundo, R.~Desai, T.~Zhu, S.~Parajuli, M.~Guo, et~al.
\newblock The many faces of robustness: A critical analysis of out-of-distribution generalization.
\newblock In {\em Proceedings of the IEEE/CVF international conference on computer vision}, pages 8340--8349, 2021.

\bibitem{hendrycks2020pretrained}
D.~Hendrycks, X.~Liu, E.~Wallace, A.~Dziedzic, R.~Krishnan, and D.~Song.
\newblock Pretrained transformers improve out-of-distribution robustness.
\newblock {\em arXiv preprint arXiv:2004.06100}, 2020.

\bibitem{sgd_cls}
M.~A. Hussain and L.~Gogoi.
\newblock Performance analyses of five neural network classifiers on nodule classification in lung ct images using weka: a comparative study.
\newblock {\em Physical and Engineering Sciences in Medicine}, 45(4):1193--1204, 2022.

\bibitem{autism}
P.~Hutson and J.~Hutson.
\newblock Autism in females: Understanding the overlooked diagnoses, unique challenges, and recommendations, 2023.

\bibitem{Jensen2003PainScales}
M.~P. Jensen, P.~Karoly, and S.~Braver.
\newblock Interpretation of visual analog scale ratings and change scores: A reanalysis of two clinical trials of postoperative pain.
\newblock {\em The Journal of Pain}, 4(7):407--414, 2003.

\bibitem{hsv}
G.~Jeon.
\newblock Color image enhancement by histogram equalization in heterogeneous color space.
\newblock {\em Int. J. Multimedia Ubiquitous Eng}, 9(7):309--318, 2014.

\bibitem{kleinberg2016inherent}
J.~Kleinberg, S.~Mullainathan, and M.~Raghavan.
\newblock Inherent trade-offs in the fair determination of risk scores.
\newblock {\em arXiv preprint arXiv:1609.05807}, 2016.

\bibitem{sgd_svm}
P.~Kumar, S.~Happy, and A.~Routray.
\newblock A real-time robust facial expression recognition system using hog features.
\newblock In {\em 2016 International Conference on Computing, Analytics and Security Trends (CAST)}, pages 289--293. IEEE, 2016.

\bibitem{kuzucu2024uncertainty}
S.~Kuzucu, J.~Cheong, H.~Gunes, and S.~Kalkan.
\newblock Uncertainty as a fairness measure.
\newblock {\em Journal of Artificial Intelligence Research}, 81:307--335, 2024.

\bibitem{kwok2025machine}
A.~M.~H. Kwok, J.~Cheong, S.~Kalkan, and H.~Gunes.
\newblock Machine learning fairness for depression detection using eeg data.
\newblock {\em arXiv preprint arXiv:2501.18192}, 2025.

\bibitem{LeCun2015DeepLearning}
Y.~LeCun, Y.~Bengio, and G.~Hinton.
\newblock Deep learning.
\newblock {\em Nature}, 521, 2015.

\bibitem{CNN_Depression}
M.~Li, Y.~Wang, C.~Yang, Z.~Lu, and J.~Chen.
\newblock Automatic diagnosis of depression based on facial expression information and deep convolutional neural network.
\newblock {\em IEEE Transactions on Computational Social Systems}, 11(5):5728--5739, 2024.

\bibitem{lopez2017personalized}
D.~Lopez~Martinez, R.~Picard, et~al.
\newblock Personalized automatic estimation of self-reported pain intensity from facial expressions.
\newblock In {\em Proceedings of the IEEE conference on computer vision and pattern recognition workshops}, pages 70--79, 2017.

\bibitem{loshchilov2017decoupled}
I.~Loshchilov and F.~Hutter.
\newblock Decoupled weight decay regularization.
\newblock {\em arXiv preprint arXiv:1711.05101}, 2017.

\bibitem{UNBC_Pain}
P.~Lucey, J.~F. Cohn, K.~M. Prkachin, P.~E. Solomon, and I.~Matthews.
\newblock Painful data: The unbc-mcmaster shoulder pain expression archive database.
\newblock In {\em 2011 IEEE International Conference on Automatic Face \& Gesture Recognition (FG)}, pages 57--64, 2011.

\bibitem{mehrabi2021survey}
N.~Mehrabi, F.~Morstatter, N.~Saxena, K.~Lerman, and A.~Galstyan.
\newblock A survey on bias and fairness in machine learning.
\newblock {\em ACM computing surveys (CSUR)}, 54(6):1--35, 2021.

\bibitem{micikevicius2018mixed}
P.~Micikevicius, S.~Narang, J.~Alben, G.~Diamos, E.~Elsen, D.~Garcia, B.~Ginsburg, M.~Houston, O.~Kuchaiev, G.~Venkatesh, et~al.
\newblock Mixed precision training.
\newblock {\em arXiv preprint arXiv:1710.03740}, 2018.

\bibitem{muraina2022ideal}
I.~Muraina.
\newblock Ideal dataset splitting ratios in machine learning algorithms: general concerns for data scientists and data analysts.
\newblock In {\em 7th international Mardin Artuklu scientific research conference}, pages 496--504, 2022.

\bibitem{rocauc}
S.~Narkhede.
\newblock Understanding auc-roc curve.
\newblock {\em Towards data science}, 26(1):220--227, 2018.

\bibitem{observer-pain-scale}
S.~Nerella, Z.~Guan, A.~Davidson, Y.~Ren, T.~Baslanti, B.~Armfield, P.~Tighe, A.~Bihorac, and P.~Rashidi.
\newblock Detecting visual cues in the intensive care unit and association with patient clinical status.
\newblock {\em arXiv preprint arXiv:2311.00565}, 2023.

\bibitem{scikit-learn}
F.~Pedregosa, G.~Varoquaux, A.~Gramfort, V.~Michel, B.~Thirion, O.~Grisel, M.~Blondel, P.~Prettenhofer, R.~Weiss, V.~Dubourg, J.~Vanderplas, A.~Passos, D.~Cournapeau, M.~Brucher, M.~Perrot, and E.~Duchesnay.
\newblock Scikit-learn: Machine learning in {P}ython.
\newblock {\em Journal of Machine Learning Research}, 12:2825--2830, 2011.

\bibitem{MLFairness}
D.~Pessach and E.~Shmueli.
\newblock A review on fairness in machine learning.
\newblock {\em ACM Computing Surveys (CSUR)}, 55(3):1--44, 2022.

\bibitem{prkachin2008structure}
K.~M. Prkachin and P.~E. Solomon.
\newblock The structure, reliability and validity of pain expression: Evidence from patients with shoulder pain.
\newblock {\em Pain}, 139(2):267--274, 2008.

\bibitem{ViTvsCNN}
M.~Rodrigo, C.~Cuevas, and N.~Garc{\'\i}a.
\newblock Comprehensive comparison between vision transformers and convolutional neural networks for face recognition tasks.
\newblock {\em Scientific Reports}, 14(1):21392, 2024.

\bibitem{ml-paindetection}
S.~D. Roy, M.~K. Bhowmik, P.~Saha, and A.~K. Ghosh.
\newblock An approach for automatic pain detection through facial expression.
\newblock {\em Procedia Computer Science}, 84:99--106, 2016.

\bibitem{parametergrid}
M.~Salama.
\newblock Optimization of regression models using machine learning: A comprehensive study with scikit-learn.
\newblock {\em Optimization of Regression Models Using Machine Learning: A Comprehensive Study with Scikit-learn| IUSRJ}, 5, 2024.

\bibitem{brave_men}
A.~Samulowitz, I.~Gremyr, E.~Eriksson, and G.~Hensing.
\newblock “brave men” and “emotional women”: A theory-guided literature review on gender bias in health care and gendered norms towards patients with chronic pain.
\newblock {\em Pain research and management}, 2018(1):6358624, 2018.

\bibitem{saqib2021machine}
K.~Saqib, A.~F. Khan, and Z.~A. Butt.
\newblock Machine learning methods for predicting postpartum depression: scoping review.
\newblock {\em JMIR mental health}, 8(11):e29838, 2021.

\bibitem{Scholkopf2002SVM}
B.~Sch{\"o}lkopf.
\newblock Learning with kernels: support vector machines, regularization, optimization, and beyond, 2002.

\bibitem{rotation}
C.~Shorten and T.~M. Khoshgoftaar.
\newblock A survey on image data augmentation for deep learning.
\newblock {\em Journal of big data}, 6(1):1--48, 2019.

\bibitem{ResNetCNN}
Y.~Shuang, G.~Liangbo, Z.~Huiwen, L.~Jing, C.~Xiaoying, S.~Siyi, Z.~Xiaoya, and L.~Wen.
\newblock Classification of pain expression images in elderly with hip fractures based on improved resnet50 network.
\newblock {\em Frontiers in Medicine}, 11, 2024.

\bibitem{spitale2024underneath}
M.~Spitale, J.~Cheong, and H.~Gunes.
\newblock Underneath the numbers: Quantitative and qualitative gender fairness in llms for depression prediction.
\newblock {\em arXiv preprint arXiv:2406.08183}, 2024.

\bibitem{sun2020evolution}
W.~Sun, O.~Nasraoui, and P.~Shafto.
\newblock Evolution and impact of bias in human and machine learning algorithm interaction.
\newblock {\em Plos one}, 15(8):e0235502, 2020.

\bibitem{balance_trick}
S.~Susan and A.~Kumar.
\newblock The balancing trick: Optimized sampling of imbalanced datasets—a brief survey of the recent state of the art.
\newblock {\em Engineering Reports}, 3(4):e12298, 2021.

\bibitem{imbalance}
S.~Tyagi and S.~Mittal.
\newblock Sampling approaches for imbalanced data classification problem in machine learning.
\newblock In {\em Proceedings of ICRIC 2019: Recent innovations in computing}, pages 209--221. Springer, 2020.

\bibitem{varona2022discrimination}
D.~Varona and J.~L. Su{\'a}rez.
\newblock Discrimination, bias, fairness, and trustworthy ai.
\newblock {\em Applied Sciences}, 12(12):5826, 2022.

\bibitem{vashney2022trustworthy}
K.~R. Vashney.
\newblock {\em Trustworthy machine learning}.
\newblock Independently published, 2022.

\bibitem{bias}
M.~B. Vela, A.~I. Erondu, N.~A. Smith, M.~E. Peek, J.~N. Woodruff, and M.~H. Chin.
\newblock Eliminating explicit and implicit biases in health care: evidence and research needs.
\newblock {\em Annual review of public health}, 43(1):477--501, 2022.

\bibitem{fairnessdefinitions}
S.~Verma and J.~Rubin.
\newblock Fairness definitions explained.
\newblock In {\em Proceedings of the international workshop on software fairness}, pages 1--7, 2018.

\bibitem{wang2023bias}
R.~Wang, P.~Chaudhari, and C.~Davatzikos.
\newblock Bias in machine learning models can be significantly mitigated by careful training: Evidence from neuroimaging studies.
\newblock {\em Proceedings of the National Academy of Sciences}, 120(6):e2211613120, 2023.

\bibitem{werner2019automatic}
P.~Werner, D.~Lopez-Martinez, S.~Walter, A.~Al-Hamadi, S.~Gruss, and R.~W. Picard.
\newblock Automatic recognition methods supporting pain assessment: A survey.
\newblock {\em IEEE Transactions on Affective Computing}, 13(1):530--552, 2019.

\bibitem{logreg-paindetection}
B.~D. Winslow, R.~Kwasinski, K.~Whirlow, E.~Mills, J.~Hullfish, and M.~Carroll.
\newblock Automatic detection of pain using machine learning.
\newblock {\em Frontiers in pain research}, 3:1044518, 2022.

\bibitem{intensity_hist_equ}
C.~Y. Wong, G.~Jiang, M.~A. Rahman, S.~Liu, S.~C.-F. Lin, N.~Kwok, H.~Shi, Y.-H. Yu, and T.~Wu.
\newblock Histogram equalization and optimal profile compression based approach for colour image enhancement.
\newblock {\em Journal of Visual Communication and Image Representation}, 38:802--813, 2016.

\bibitem{xu2020investigating}
T.~Xu, J.~White, S.~Kalkan, and H.~Gunes.
\newblock Investigating bias and fairness in facial expression recognition.
\newblock In {\em Computer Vision--ECCV 2020 Workshops: Glasgow, UK, August 23--28, 2020, Proceedings, Part VI 16}, pages 506--523. Springer, 2020.

\bibitem{fairness-aware}
J.~Yang, A.~A. Soltan, D.~W. Eyre, Y.~Yang, and D.~A. Clifton.
\newblock An adversarial training framework for mitigating algorithmic biases in clinical machine learning.
\newblock {\em NPJ digital medicine}, 6(1):55, 2023.

\bibitem{nystroem}
T.~Yang, Y.-F. Li, M.~Mahdavi, R.~Jin, and Z.-H. Zhou.
\newblock Nystr{\"o}m method vs random fourier features: A theoretical and empirical comparison.
\newblock {\em Advances in neural information processing systems}, 25, 2012.

\bibitem{disparateimpact}
M.~B. Zafar, I.~Valera, M.~Gomez~Rodriguez, and K.~P. Gummadi.
\newblock Fairness beyond disparate treatment \& disparate impact: Learning classification without disparate mistreatment.
\newblock In {\em Proceedings of the 26th international conference on world wide web}, pages 1171--1180, 2017.

\bibitem{adversarial-debiasing}
B.~H. Zhang, B.~Lemoine, and M.~Mitchell.
\newblock Mitigating unwanted biases with adversarial learning.
\newblock In {\em Proceedings of the 2018 AAAI/ACM Conference on AI, Ethics, and Society}, pages 335--340, 2018.

\bibitem{auc_eq}
X.~Zhang, X.~Li, Y.~Feng, and Z.~Liu.
\newblock The use of roc and auc in the validation of objective image fusion evaluation metrics.
\newblock {\em Signal processing}, 115:38--48, 2015.

\end{thebibliography}

\end{document}